\documentclass[journal,twoside,web]{ieeecolor}
\usepackage{generic}
\usepackage{cite}
\usepackage{amsmath,amssymb,amsfonts}
\usepackage{graphicx}
\usepackage{hyperref}
\hypersetup{hidelinks=true}
\usepackage{textcomp}

\usepackage{multirow}
\graphicspath{{figures/}}

\usepackage{dsfont}
\usepackage{mathrsfs}
\usepackage{caption}
\usepackage{subcaption}
\usepackage{subfloat}
\usepackage{multicol}
\usepackage{diagbox}
\usepackage{comment}
\usepackage[dvipsnames]{xcolor, colortbl}

\usepackage[noend]{algorithm, algpseudocode}
\floatname{algorithm}{Algorithm}

\newcommand\blfootnote[1]{%
  \begingroup
  \renewcommand\thefootnote{}\footnote{#1}%
  \addtocounter{footnote}{-1}%
  \endgroup
}

\algnewcommand{\LeftComment}[1]{\Statex \(\triangleright\) #1}
\DeclareFontFamily{U}{mathx}{\hyphenchar\font45}
\DeclareFontShape{U}{mathx}{m}{n}{
      <5> <6> <7> <8> <9> <10>
      <10.95> <12> <14.4> <17.28> <20.74> <24.88>
      mathx10
      }{}
\DeclareSymbolFont{mathx}{U}{mathx}{m}{n}
\DeclareFontSubstitution{U}{mathx}{m}{n}
\DeclareMathAccent{\widebar}{0}{mathx}{"73}
\def\BibTeX{{\rm B\kern-.05em{\sc i\kern-.025em b}\kern-.08em
    T\kern-.1667em\lower.7ex\hbox{E}\kern-.125emX}}
\providecommand{\refname}{REFERENCES}

\begin{document}
\title{To do(\(x\)) or not to do(\(x\)): Medical Image Counterfactuals for Dataset Augmentation}
\author{Yasin Ibrahim, Robin J. Evans, and Konstantinos Kamnitsas
}

\maketitle

\begin{abstract}
Medical image analysis is often hindered by biased datasets, which can lead to biased models and limited clinical applicability. A promising strategy for mitigating such biases is to augment training data with synthetic images. Counterfactual (CF) generation is one such strategy, though the term is used in two different senses: in some works, CFs are produced through causality-based interventions derived from structural causal models, whereas in others, they are produced by non-causal image edits or conventional conditional generative models, such as altering anatomy or adding pathologies. In this work, we study this distinction and evaluate its practical consequences for medical image augmentation. We compare three conditioning strategies: \textit{Deterministic}, which changes selected variables while holding the remaining variables fixed; \textit{Undirected}, which updates variables according to learned statistical associations without assigning causal directions; and \textit{Causal}, which propagates interventions along a directed causal graph. We analyse how these choices affect the resulting images, and explore when causally grounded methods improve dataset augmentation or bring limited benefit. In particular, we assess downstream performance and fairness, where fairness refers to reduced sensitivity to dataset biases across sensitive subgroups. Our experiments demonstrate that using a causal approach to synthetic training data generation can lead to tangible benefits, with these insights offering valuable guidance  to machine learning practitioners for the effective design of data generation protocols.

\end{abstract}


\blfootnote{This work was sumbitted on 11 September 2026. Yasin Ibrahim (yasin.ibrahim@eng.ox.ac.uk) and Konstantinos Kamnitsas (konstantinos.kamnitsas@eng.ox.ac.uk) are with the Department of Engineering Science, University of Oxford. Robin J. Evans (evans@stats.ox.ac.uk) is with the Department of Statistics, University of Oxford. Yasin Ibrahim is supported by the EPSRC Centre for Doctoral Training in Health Data Science (EP/S02428X/1).}

\section{Introduction}

Medical image analysis faces the challenge of biased datasets, which can lead to biased models \cite{bias} and hinder the deployment of such models in real-world clinical scenarios \cite{dlchallenge, nofairlunch}. One way to address this challenge is to generate synthetic data to create more balanced training datasets and counteract undesirable correlations \cite{biasgenerate}. Indeed, various works have shown that augmenting training datasets with synthetic data can improve the performance and subgroup fairness of downstream models, such as disease predictors for chest x-ray, skin lesion, and histopathology data \cite{googlefair}, breast cancer MRI \cite{augmentcfdiffusion}, and text-prompted generation of skin lesion images \cite{augmentdiffusion}. Fully synthetic training sets have also been explored, although these generally do not perform as well as augmenting real datasets with synthetic samples \cite{syntheticenhanced}. However, synthetic data is not automatically beneficial: if the generation process reproduces or strengthens biased correlations present in the training set, downstream models may inherit or amplify these biases \cite{llmsyntheticbias}. This concern is supported by theoretical analyses \cite{mids, biasamptheory} and empirical evidence \cite{deepgenbias}. These findings motivate a careful examination of how synthetic medical images are generated and used for augmentation. In recent years, there has been growing interest in integrating causality into machine learning models \cite{intro, Peters17}. Medical imaging is an application area of particular promise, with several works arguing that causal reasoning can help address confounding, dataset shift, and robustness issues in image analysis \cite{causalitymatters, causalmedreview}. This interest has been especially prominent in the generation of counterfactuals (CFs). In the causal sense, CF generation involves specifying a causal diagram over relevant variables, estimating how an intervention on one variable should affect the others, and then using these inferred variable values to generate a corresponding image \cite{deepscm, neuralscm1}. However, there remains limited systematic evidence on when such causally grounded mechanisms provide practical benefits such as bias mitigation, rather than simpler conditional or deterministic forms of generation.

Alongside its formal definition in the causality literature, the term ``counterfactual'' is also commonly used in medical imaging to describe synthetic modifications to an input image, such as increasing ventricular size in a brain MRI scan or adding signs of disease to a chest x-ray \cite{fakecf1}. Such modifications may be produced by direct image processing procedures or by conventional generative models that edit an image conditioned on a desired attribute, without explicitly modelling causal relations among the associated variables. Because synthetic data generation in medicine is sensitive to both clinical validity and downstream bias, this distinction matters. Therefore, in this work, we distinguish between \textit{causal counterfactuals}, which are derived from interventions in a causal model, and broader \textit{image counterfactuals}, which modify images without necessarily encoding causal relations. This distinction allows different methods to be evaluated and compared according to the assumptions they actually make. To study the consequences of these assumptions, we compare three conditioning strategies for generative models. In the \textit{Deterministic} approach, given an input image and associated variables, one or more variables are set to specified values while all remaining variables are kept fixed. In the \textit{Undirected} approach, relationships between variables are modelled statistically, so changing one variable can affect other associated variables, but the directions of these relationships are not treated as causal. In the \textit{Causal} approach, relationships are represented by a directed causal graph, so interventions propagate only through the causal directions encoded in the graph. We first investigate how the output counterfactuals produced by these methods differ depending on the assumed graph and the variables altered. We then explore using data generated by these different methods for training dataset augmentation and the effects this has on the accuracy and fairness, with regards to bias mitigation, of a predictor trained on this. We divide the actual data generation approach into two methods: counterfactuals of real images from the training set, and purely synthetic images sampled from the latent space of the generative model. From this, we gain insights into when it is beneficial to use an approach that explicitly considers causal relations. In summary, we make the following contributions:

\begin{itemize}
    \item We demarcate a clear distinction between counterfactual methods that do and do not use causal inference, and provide a categorisation of dataset augmentation approaches.
    \item We evaluate the effects of conditioning generative models with techniques that impose different assumptions about which relationships between modelled variables should be preserved after an intervention.
    \item We identify the conditions under which generating data using explicitly causal methods for training dataset augmentation can improve downstream performance and fairness over non-causal alternatives.
\end{itemize}

\section{Related Work}

To mitigate the issue of bias in medical imaging \cite{bias, dlchallenge, nofairlunch}, a range of strategies has been proposed, including reweighting or resampling-based objectives \cite{zong2023medfair} and domain adaptation methods for distribution shift \cite{li2023medical}. However, what has been gaining particular traction is the use of data augmentation techniques to rebalance datasets and improve the coverage of underrepresented samples \cite{goceri2023medical, biasgenerate}. By synthesising realistic medical images, researchers can augment datasets with specific demographic groups, pathologies, acquisition conditions, or imaging appearances that are underrepresented in the real data distribution. Such methods are especially relevant when collecting additional real medical imaging data is expensive and time-consuming, or when data sharing is constrained by privacy, regulatory, and governance requirements \cite{guibas2017synthetic}.

Generative modelling techniques have played a central role in the development of synthetic medical data, with some of the most promising results have been achieved with Generative Adversarial Networks (GANs) \cite{gan}, which have been used for tasks such as lesion synthesis, MRI reconstruction, and modality translation \cite{ganmed1, ganmed2}. Other deep generative approaches, including variational models \cite{vae}, normalising-flow-based models \cite{normalisingflows}, and autoregressive or discrete-latent models, have also been explored for medical image synthesis and representation learning. More recently, diffusion-based generative models have shown strong performance in producing high-fidelity medical images while maintaining anatomical consistency \cite{diffusionmed1}. Synthetic generation has also been used directly for bias mitigation and fairness-oriented augmentation \cite{googlefair}. Though these methods are powerful, they often learn observational correlations from the training data, so when the training distribution contains confounding or undesirable correlations, generated samples may preserve or amplify these patterns unless the generation process is carefully controlled.

Causal reasoning offers one way to make such control more explicit. In causal medical image modelling, relationships between variables are represented not only as statistical associations but as directed cause-effect assumptions \cite{causalitymatters, causalmedreview}. This perspective is particularly relevant for counterfactual generation; in a causal counterfactual, an intervention such as changing disease status, age, or acquisition parameters is propagated through a structural causal model, so only variables downstream of the intervention should change, while other variables should remain fixed except through the specified causal pathways \cite{deepscm, neuralscm1}. This differs from conventional undirected generation, where variables may be associated but the model does not distinguish whether the association corresponds to a causal effect, a reverse effect, or a common cause. Causal counterfactuals therefore provide a principled framework for specifying which aspects of an image should and should not change under an intervention.

By contrast, many works in medical imaging use the term counterfactual more broadly to refer to any synthetic modification that changes a visual attribute while preserving sample identity \cite{fakecf1}. Examples include increasing lesion size, introducing synthetic pathologies, modifying anatomical structures, or translating an image to a target domain defined by a desired attribute. These approaches are valuable for data augmentation, visual explanation, and stress testing, but they often lack explicit causal grounding: the modification is not necessarily derived from a model of how the underlying variables are causally related. This difference is important because non-causal image edits may produce realistic-looking samples while still changing variables that should have remained fixed, or preserving variables that should have changed under a causal intervention.

In this paper, we categorise methods that synthesise counterfactual-style medical images into three families. \textit{Deterministic} methods set selected variables to desired values and keep all other conditioning variables fixed. These include many direct image-editing or image-to-image translation approaches, such as methods based on CycleGAN \cite{cyclegan} or Pix2Pix \cite{pix2pix}, when they are used to map an image to a target domain without modelling how the remaining variables should change. \textit{Undirected} methods model statistical relationships among variables and use these associations when generating or editing images, but do not assign causal directions to those relationships. \textit{Causal} methods explicitly encode directed dependencies between variables, often through Structural Causal Models or Neural Causal Models \cite{deepscm, neuralscm1}, so that interventions propagate through the graph in a way that is consistent with the assumed causal structure. Our contribution is to compare these families under a common experimental framework and evaluate when the stronger assumptions of causal conditioning translate into measurable benefits for dataset augmentation, downstream performance, and fairness.

\section{Background}
\label{sec:background}

A Structural Causal Model (SCM) is defined as a tuple \cite{neuralscm1}:
\[
    \mathcal{M} = \langle V,U,\mathcal{F},P(U)\rangle,
\]
where \(V=\{v_1,\dots,v_n\}\) denotes the endogenous variables, \(U=\{u_1,\dots,u_n\}\) the exogenous noise variables, \(P(U)\) their distribution, and \(\mathcal{F}=\{f_1,\dots,f_n\}\) the structural mechanisms. Each endogenous variable is determined by its parents \(\text{pa}_i\) and corresponding noise variable \(u_i\):
\begin{equation}
    v_i := f_i(\text{pa}_i,u_i).
    \label{eq:func}
\end{equation}

Unlike purely observational models, SCMs also support interventions through the \(\text{do}\)-operator \cite{foundation}. Applying \(\text{do}(v_i=a)\) fixes \(v_i\) to \(a\) and removes the influence of its parents, resulting in
\[
    P(V|\text{do}(v_i=a))
    =
    \prod_{j\neq i}p(v_j|pa_j)\cdot\mathds{1}_{\{v_i=a\}}.
\]

To obtain sample-specific counterfactuals rather than population-level intervention effects, SCMs use the abduction--action--prediction procedure \cite{Peters17}:
\begin{enumerate}
    \item \textbf{Abduction}: Infer the sample-specific exogenous noise through \(p(U|V)\).
    \item \textbf{Action}: Apply \(\text{do}(V=A)\) to obtain the modified model \(\mathcal{\widetilde{M}}_{\text{do}(V=A)}\).
    \item \textbf{Prediction}: Use the modified model to estimate the counterfactual variables.
\end{enumerate}

Various methods have been proposed to apply this to the problem of generating image counterfactuals; these include diffusion \cite{sanchezdiffusion}, VAE \cite{deepscm}, hierarchical VAE \cite{hvae} and semi-supervised \cite{ibrahim2024semi} approaches. In this paper, we will use \(x\) to denote the image and \textbf{pa} to denote the parent variables, such as age, sex and race for chest x-rays, which ``cause'' the image appearance. The counterfactual parents (post-intervention) are denoted by \(\widetilde{\text{\textbf{pa}}}\).

\section{Methods}
\subsection{Model Conditioning}
\label{sec:modelcond}

In the medical imaging literature, the term ``counterfactual'' is often used to describe any change to an image that maintains the sample identity, such as the removal or addition of lesions from a scan, via direct image processing or a (non-causal) undirected generative model. An alternative interpretation of ``counterfactual,'' however, is the causal approach described in Section \ref{sec:background} applied to the relevant variables in the model, such as age, disease, etc. In the context of image modification, this is typically the initial step before the inferred variable values are input to a conditional generative model, which then generates the corresponding altered image. To avoid further confusion we employ the terms \textbf{image counterfactual} and \textbf{causal counterfactual} to refer to these two concepts respectively.

\begin{figure}[h]
\begin{subfigure}[c]{0.33\linewidth}
\centering
    \includegraphics[width=0.85\linewidth]{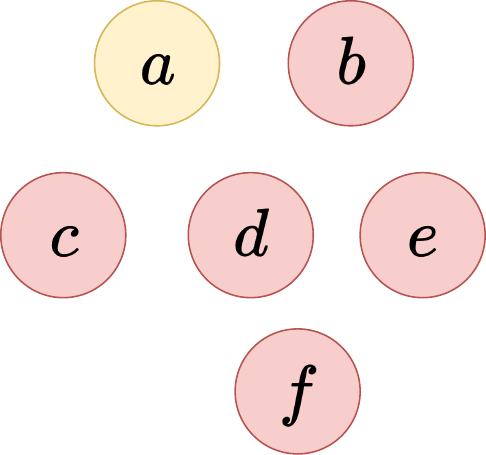}  \caption{\label{fig:deteg}Deterministic}
\end{subfigure}\hfill
\begin{subfigure}[c]{0.33\linewidth}
  \centering
    \includegraphics[width=0.85\linewidth]{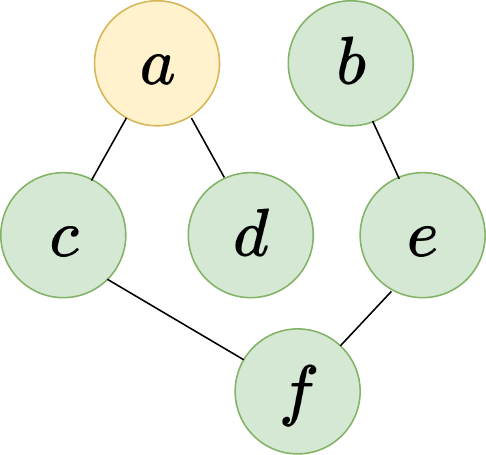}
  \caption{\label{fig:condeg}Undirected}
\end{subfigure}\hfill
\begin{subfigure}[c]{0.33\linewidth}
  \centering
    \includegraphics[width=0.85\linewidth]{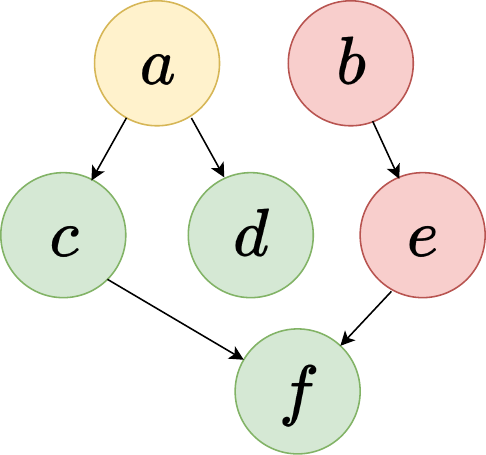}
  \caption{\label{fig:causaleg}Causal}
\end{subfigure}
\caption{Effect of intervention with each approach; Yellow: intervened, Green: affected, Red: unaffected.}
\label{fig:egs}
\end{figure}

We thus seek to identify when the additional cost of generating causal counterfactuals is beneficial. To investigate this, we propose demarcating three distinct methods for conditioning generative models with multiple variables and relationships between them. Fig.~\ref{fig:egs} provides an example of the differences between them; we have a set of variables \((a)-(f)\) with a specified causal structure, and intervene on the first of these via \(a=\tilde{a}\). The three conditioning approaches are:

\begin{enumerate}
    \item \textbf{Deterministic}: The relationships between variables are ignored. We set the desired subset of variables to new values and force the remainder to retain their original values. For the example in Fig.~\ref{fig:deteg}, we have:
    \[p(\widetilde{\text{\textbf{pa}}}) = \mathds{1}_{\{a=\tilde{a}\}} \cdot \prod_{v\neq a}\mathds{1}_{\{v=v\}}\]
    This is the most common approach for conditioning standard (non-causal) generative models.
    \item \textbf{Undirected}: Assuming we know which variables are associated, we impose an undirected graph structure between them. We set a desired subset of variables to new values and infer the remainder using the joint distribution. For the example in Fig.~\ref{fig:condeg}, the post-intervention distribution becomes:
    \[p(\widetilde{\text{\textbf{pa}}}) = \mathds{1}_{\{a=\tilde{a}\}} \cdot p(b,c,d,e,f|a=\tilde{a})\]
    This is a stronger prior than the previous case, as it assumes we know which variables are related.
    
    \item \textbf{Causal}: Assuming we know the causal relations between variables, we impose a directed graph structure so that child variables are only affected by their parents. We conduct the abduction-action-prediction procedure (Section \ref{sec:background}) to create CFs. As in Fig.~\ref{fig:causaleg}, inference is made via the model \(\widetilde{\mathcal{M}}_{\text{do}(a=\tilde{a})}\). This is an even stronger prior, with assumed known direction of causality, where only d-connected variables are affected by the intervention.
    
\end{enumerate}

\subsection{Image Generation}
\label{sec:img_generation}

After training, we consider two approaches to generate synthetic images for dataset augmentation. Alongside the counterfactual paradigm described above, we also explore generating new images by sampling directly from the latent space (i.e. ``de novo'' sampling, not conditioned on input image). Such an approach has been shown to improve performance of downstream classifiers \cite{googlefair}. So far, however, this has been limited to the case where there are no assumed relationships between variables -- \textit{Deterministic} in our terminology. To do this, for each approach, each variable is either assigned a desired value or sampled from an unconditional distribution. 

To enrich our analysis, we adapt the other two conditioning approaches to this setting. In both cases, we begin by choosing a variable and either assign it to a specific value or sample it from its marginal distribution. We then sequentially sample the remaining variables conditionally on the previously sampled ones. In the \textit{Undirected} approach, this means any connection between variables induces a conditional distribution and only variables that are entirely disconnected in the proposed graph are sampled from their marginal distributions. For \textit{Causal}, on the other hand, children variables are conditioned on their parents but the reverse is not true. As in the \textit{Deterministic} case \cite{googlefair}, these variables are then all fed to the conditional decoder of the generative model to produce a new image. This image generation paradigm is described in Algorithm~\ref{alg:samp}.\\ 

\begin{algorithm}
    \caption{Sampling}
    \begin{algorithmic}[1]
    \Require Parents (\(a\))
    \Ensure Generated Image (\(\bar{x}\))
    \State \(a \sim p(a)\) \Comment Sample or set one of the variables
    \State \(\widebar{\text{\textbf{pa}}} \sim p(\text{\textbf{pa}}|a=\tilde{a})\) \Comment Sample the remaining variables
    \State \(z \sim p(z)\) \Comment Sample a latent \(z\) from the prior of the generative model
    \State \(\bar{x} = \text{dec}(z, \widebar{\text{\textbf{pa}}})\) \Comment Feed sampled vars and \(z\) to decoder
    \State \Return{\(\bar{x}\)}
    \end{algorithmic}
    \label{alg:samp}
\end{algorithm}

\begin{algorithm}
    \caption{Counterfactuals}
    \begin{algorithmic}[1]
    \Require Image (\(x\)), Counterfactual parent (\(\tilde{a}\))
    \Ensure Counterfactual Image (\(\tilde{x}\))
    
    \State \(z\) = enc(\(x\)) \Comment Encode \(x\) to obtain latent representation \(z\) 
    \State \(\widehat{\text{\textbf{pa}}} = f(x)\) \Comment Predict missing variables from image \(x\)
    
    \State \(\widetilde{\text{\textbf{pa}}} \leftarrow (a = \tilde{a})\) \Comment Intervene on one of the variables
    
    \State \(\widetilde{\text{\textbf{pa}}} \sim \widetilde{\mathcal{M}}_{\text{do}(a=\tilde{a})}\) \Comment Infer the remaining variables
    
    \State \(\tilde{x} = \text{dec}(z, \widetilde{\text{\textbf{pa}}})\)  \Comment Feed altered vars and \(z\) to decoder
    
    \State \Return{\(\tilde{x}\)}
    \end{algorithmic}
    \label{alg:cf}
\end{algorithm}

\subsection{Architecture}
\begin{figure}
    \centering
    \includegraphics[width=0.5\textwidth]{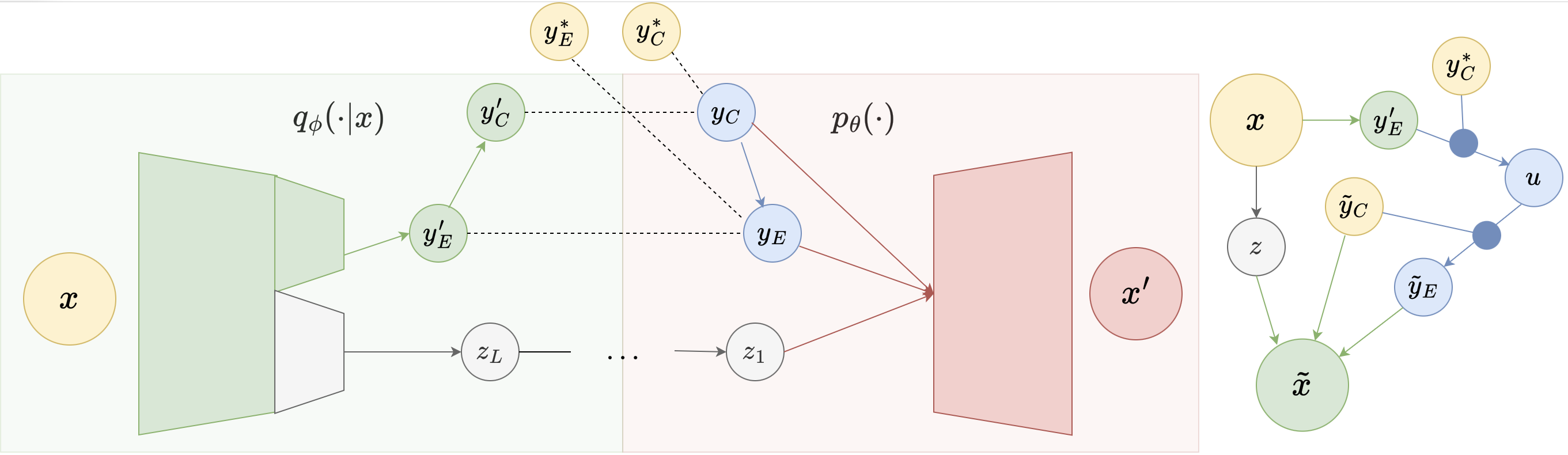}
    \caption{Model outline. Green: observed, Grey: latent, Red: predicted, Blue: causal generative, Yellow: decoding. (left) Training: we use the \(y\) predictions for decoding unless they are observed, (right) CF generation.}
    \label{fig:model}
\end{figure}
We use the semi-supervised causal generative model proposed by \cite{ibrahim2024semi} as this allows us to explore the conditioning approaches in greater depth; we can alter the amount of available labelled data, e.g. by pretending that a percentage of values of endogeneous variables are missing, and observe the effect this has on the outputs of each approach (see Section \ref{subsec:cxr2}). By adapting the conditioning module, we implement each of the three approaches outlined in Section \ref{sec:modelcond}.

Firstly, for the \textit{Causal} approach, we use \cite{ibrahim2024semi} directly. This involves a hierarchical VAE that encodes the input image to multiple scales before decoding it at multiple latent levels. A DAG is also specified and the arrows (relationships between variables) are learned as normalising flows \cite{normalisingflows}. In addition, a variable predictor is trained as an extra head on the encoder, used to predict any missing values.

For the \textit{Undirected} approach, we keep the DAG but add functions (again implemented as normalising flows) in the opposite direction to the existing arrows, to form bi-directed edges and model undirected relations. For example, for DAG a) in Fig.~\ref{fig:mnistdags}, we also model the function \(p(d|b,f)\). The model is trained as the original \cite{ibrahim2024semi} with the added flows also learned and used to infer any missing values. To produce counterfactuals, we set the desired variable values and infer the remaining using the learned conditional dependencies. Finally, for the \textit{Deterministic} approach, we eschew the causal graph entirely and use only the encoder from the generative model to predict missing values. Then, to produce counterfactuals, we infer the latent \(z\) and input the desired variable values as conditions.

\section{Analysis on Morpho-MNIST}
\label{sec:mnist}
We now explore the approaches described in Section \ref{sec:modelcond} when different causal relationships govern the data generating process to study when these causal modelling assumptions are useful.
To begin, we use Morpho-MNIST \cite{morphomnist}, which provides a framework for generating data with causal relationships between variables of our own specification, with a known generative process and distribution, and thus we can evaluate counterfactuals from learned generative models against the ground-truth log-likelihoods. This is impossible with real medical data, where the true underlying distribution is unknown. These experiments therefore provide insights that guide and complement further exploration on medical data.

\begin{figure}[t!]
\centering
    \subfloat{%
        \includegraphics[width=.15\linewidth]{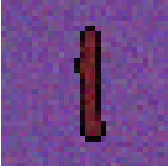}%
    }
    \subfloat{%
        \includegraphics[width=.15\linewidth]{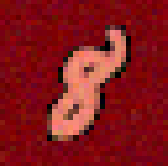}%
    }
    \subfloat{%
        \includegraphics[width=.15\linewidth]{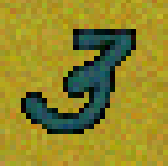}%
    }
    \subfloat{%
        \includegraphics[width=.15\linewidth]{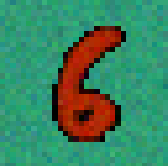}%
    }
    \subfloat{%
        \includegraphics[width=.15\linewidth]{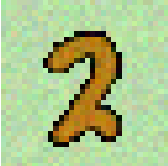}%
    }
    \subfloat{%
        \includegraphics[width=.15\linewidth]{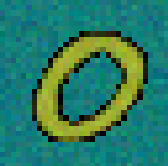}%
    }
    \caption{Example coloured MNIST images}
    \label{fig:cmmnist}
\end{figure}

\begin{figure*}[h]
    \centering
    \includegraphics[width=\textwidth]{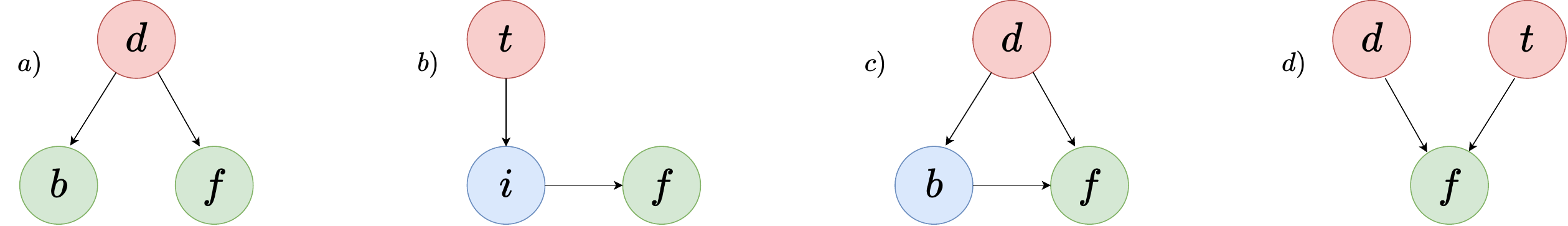}
    \caption{We implemented four generative processes for synthesising four coloured MNIST datasets, when relations between variables are follow the above DAGs. Variables: Digit $d$, thickness $t$, intensity $i$, background colour $b$, foreground colour $f$.}
    \label{fig:mnistdags}

    \vspace{0.8em}

    \begin{minipage}[t]{0.245\textwidth} 
        \centering
        \scalebox{0.64}{
        \begin{tabular}{c|c|c|c|c}
        Int. & Effect & Det. & Undir. & Causal \\
        \hline
        \multirow{2}{*}{\textcolor{WildStrawberry}{$d$}} & \textcolor{OliveGreen}{$f$} (\(\uparrow\)) & -2.56 & \underline{-1.52} & \textbf{-1.51} \\
        & \textcolor{OliveGreen}{$b$} (\(\uparrow\)) & -2.55 & \underline{-1.75} & \textbf{-1.72} \\
        \hline
        \multirow{2}{*}{\textcolor{OliveGreen}{$f$}} & \textcolor{OliveGreen}{$b$} (\(\uparrow\)) & \underline{-1.15} & -2.12 & \textbf{-1.14} \\
        & \textcolor{WildStrawberry}{$d$} (\(\downarrow\)) & \textbf{0} & 0.17 & \textbf{0} \\
        \hline
        \multirow{2}{*}{\textcolor{OliveGreen}{$b$}} & \textcolor{OliveGreen}{$f$} (\(\uparrow\)) & \textbf{-1.04} & -2.72 & \underline{-1.04} \\
        & \textcolor{WildStrawberry}{$d$} (\(\downarrow\)) & \textbf{0} & 0.16 & \textbf{0} \\
        \hline
        \multicolumn{1}{c}{} & \multicolumn{1}{c}{}  & \multicolumn{1}{c}{}  & \multicolumn{1}{c}{}  & \multicolumn{1}{c}{} \\ 
        \multicolumn{1}{c}{} & \multicolumn{1}{c}{}  & \multicolumn{1}{c}{}  & \multicolumn{1}{c}{}  & \multicolumn{1}{c}{} \\ 
        \end{tabular}
        }
        \captionof{table}{CFs on DAG (a).}
        \label{tab:donfgbg}
    \end{minipage}
    \hfill
    \begin{minipage}[t]{0.245\textwidth} 
        \centering
        \scalebox{0.64}{
        \begin{tabular}{c|c|c|c|c}
        Int. & Effect & Det. & Undir. & Causal \\
        \hline
        \multirow{2}{*}{\textcolor{WildStrawberry}{$t$}} & \textcolor{RoyalBlue}{$i$} (\(\uparrow\)) & -7.03 & \underline{-4.17} & \textbf{-4.13} \\
        & \textcolor{OliveGreen}{$f$} (\(\uparrow\)) & -5.05 & \textbf{-3.72} & \underline{-3.77} \\
        \hline
        \multirow{2}{*}{\textcolor{RoyalBlue}{$i$}} & \textcolor{WildStrawberry}{$t$} (\(\uparrow\)) & \underline{-2.19} & -5.25 & \textbf{-2.18} \\
        & \textcolor{OliveGreen}{$f$} (\(\uparrow\)) & -5.93 & \textbf{-3.91} & \underline{-3.97} \\
        \hline
        \multirow{2}{*}{\textcolor{OliveGreen}{$f$}} & \textcolor{WildStrawberry}{$t$} (\(\uparrow\)) & \underline{-2.65} & -5.27 & \textbf{-2.57} \\
        & \textcolor{RoyalBlue}{$i$} (\(\uparrow\)) & \textbf{-3.86} & -6.95 & \underline{-3.93} \\
        \hline
        \multicolumn{1}{c}{} & \multicolumn{1}{c}{}  & \multicolumn{1}{c}{}  & \multicolumn{1}{c}{}  & \multicolumn{1}{c}{} \\ 
        \multicolumn{1}{c}{} & \multicolumn{1}{c}{}  & \multicolumn{1}{c}{}  & \multicolumn{1}{c}{}  & \multicolumn{1}{c}{} \\ 
        \end{tabular}
        }
        \captionof{table}{CFs on DAG (b).}
        \label{tab:ionfg}
    \end{minipage}
    \hfill
    \begin{minipage}[t]{0.245\textwidth} 
        \centering
        \scalebox{0.64}{
        \begin{tabular}{c|c|c|c|c}
        Int. & Effect & Det. & Undir. & Causal \\
        \hline
        \multirow{2}{*}{\textcolor{WildStrawberry}{$d$}} & \textcolor{OliveGreen}{$f$} (\(\uparrow\)) & -2.73 & \textbf{-0.65} & \underline{-0.67} \\
        & \textcolor{RoyalBlue}{$b$} (\(\uparrow\)) & -2.65 & \underline{-0.43} & \textbf{-0.39} \\
        \hline
        \multirow{2}{*}{\textcolor{RoyalBlue}{$b$}} & \textcolor{OliveGreen}{$f$} (\(\uparrow\)) & -2.61 & \textbf{-0.65} & \underline{-0.66} \\
        & \textcolor{WildStrawberry}{$d$} (\(\downarrow\)) & \textbf{0} & 0.13 & \textbf{0} \\
        \hline
        \multirow{2}{*}{\textcolor{OliveGreen}{$f$}} & \textcolor{RoyalBlue}{$b$} (\(\uparrow\)) & \underline{-0.84} & -2.63 & \textbf{-0.83} \\
        & \textcolor{WildStrawberry}{$d$} (\(\downarrow\)) & \textbf{0} & 0.13 & \textbf{0} \\
        \hline
        \textcolor{WildStrawberry}{$d$} \& & \multirow{2}{*}{\textcolor{OliveGreen}{$f$} (\(\uparrow\))} & \multirow{2}{*}{-2.62} & \multirow{2}{*}{\underline{-0.99}} & \multirow{2}{*}{\textbf{-0.96}} \\
        \textcolor{RoyalBlue}{$b$} & & & & \\
        \hline
        \end{tabular}
        }
        \captionof{table}{CFs on DAG (c).}
        \label{tab:bgonfg}
    \end{minipage}
    \hfill
    \begin{minipage}[t]{0.245\textwidth} 
        \centering
        \scalebox{0.64}{
        \begin{tabular}{c|c|c|c|c}
        Int. & Effect & Det. & Undir. & Causal \\
        \hline
        \multirow{2}{*}{\textcolor{WildStrawberry}{$d$}} & \textcolor{WildStrawberry}{$t$} (\(\uparrow\)) & -5.35 & \textbf{-3.29} & \underline{-3.31} \\
        & \textcolor{OliveGreen}{$f$} (\(\uparrow\)) & -2.64 & \textbf{-1.50} & \underline{-1.54} \\
        \hline
        \multirow{2}{*}{\textcolor{WildStrawberry}{$t$}} & \textcolor{OliveGreen}{$f$} (\(\uparrow\)) & -2.24 & \underline{-1.28} & \textbf{-1.27} \\
        & \textcolor{WildStrawberry}{$d$} (\(\downarrow\)) & \textbf{0} & 0.14 & \textbf{0} \\
        \hline
        \multirow{2}{*}{\textcolor{OliveGreen}{$f$}} & \textcolor{WildStrawberry}{$t$} (\(\uparrow\)) & \underline{-2.75} & -5.30 & \textbf{-2.73} \\
        & \textcolor{WildStrawberry}{$d$} (\(\downarrow\)) & \textbf{0} & 0.16 & \textbf{0} \\
        \hline
        \textcolor{WildStrawberry}{$d$} \& & \multirow{2}{*}{\textcolor{OliveGreen}{$f$} (\(\uparrow\))} & \multirow{2}{*}{-2.42} & \multirow{2}{*}{\textbf{-1.15}} & \multirow{2}{*}{\underline{-1.19}} \\
        \textcolor{WildStrawberry}{$t$} & & & & \\
        \hline
        \end{tabular}
        }
        \captionof{table}{CFs on DAG (d).}
        \label{tab:dtonfg}
    \end{minipage}
    \caption*{TABLES I-IV: Effect of intervening on each variable in each DAG in Fig.~\ref{fig:mnistdags}. For \(f,b,t,i\) we measure log-likelihood while for \(d\) we use 1 - \(\text{softmax}_\text{correct digit}\). Arrows represent whether a higher or lower score is better. Bold denotes the best performing model, while underline denotes the second. \textit{Causal} is always the best or not statistically significantly worse than the best.}
\end{figure*}

We consider five variables to create coloured MNIST images as per \cite{ibrahim2024semi} (Fig.~\ref{fig:cmmnist}): digit \(d\), thickness \(t\), intensity \(i\), foreground colour \(f\) and background colour \(b\). We then consider four DAGs with different relationships of variables (Fig.~\ref{fig:mnistdags}) and generate a dataset according to each DAG to investigate in which settings the different conditional approaches offer benefits. For each of these 4 datasets, we train a generative model using each of the 3 conditioning approaches (\textit{Deterministic}, \textit{Undirected}, \textit{Causal}), giving 12 models in total. To evaluate the effects of interventions, we intervene on a variable and measure how  the remaining variables changed as they appear in the generative model's output image. To estimate foreground colour, background colour and digit we use external predictors $f(x)$ trained on unbiased data (variable values sampled uniformly), whereas for thickness and intensity we use tools from the Morpho-MNIST package \cite{morphomnist}. The conditional variables measured in the generated image (foreground and background colour, thickness, intensity), are evaluated using the log-likelihood under the true conditional distribution specified by the DAG. More precisely, intervention on variable \(a\) of the form \(\text{do}(a = \tilde{a})\) induces a joint distribution over the variables,
\begin{equation}
    \label{eq:cfdist}
    P(\widetilde{\text{pa}}_{\text{do}(a=\tilde{a})}) = \prod_{v \in \text{des}(a)}p(v|a = \tilde{a})
\end{equation}
where \(des(a)\) are descendants of \(a\). Suppose we seek to measure the accuracy of the generative model with regards to variable \(b\). We first generate the counterfactual image \(\tilde{x}_{(a=\tilde{a})}\) before using a pretrained variable predictor \(\phi(x)\) to obtain
$\hat{\tilde{b}} = \phi(\tilde{x}_{(a=\tilde{a})})|_b$, for which we then compute the log-likelihood against the distribution defined in \eqref{eq:cfdist}. 
The case for digit is slightly different as it is discrete, hence we assess correctness of an effect by measuring the difference in the highest softmax value of the digit variable in the model's SCM. 
Because digit is never a child variable, it does not change for \textit{Deterministic} and \textit{Causal} conditionings as these do not change parents by design, but changes only for \textit{Undirected}.

For the DAG in Fig.~\ref{fig:mnistdags}(a) that assumes one parent with two children, Table \ref{tab:donfgbg} shows that when intervening on the parent variable, digit $d$, both the \textit{Causal} and \textit{Undirected} approaches estimate the effect on the children appropriately. In contrast, the \textit{Deterministic} does not obey the causal graph so does not change these variables. Moreover, when one of the children ($f,b$) is changed, we see no effect on the parent $d$ with \textit{Causal} and \textit{Deterministic} but due to the bidirected nature of \textit{Undirected}, this change has an influence on the digit variable. These patterns are further exhibited for the other DAGs (Fig.~\ref{fig:mnistdags} b, c, d). This shows that \textit{Deterministic} keeps variables constant more than desired under the assumed DAG, while \textit{Undirected} tends to change them too readily. We gain further insights from the DAGs in Fig.~\ref{fig:mnistdags} (b) and (c) where we have variables with both a parent and a child (blue variables). Tables \ref{tab:ionfg} and \ref{tab:bgonfg} show that when we intervene on these (blue) variables, \textit{Causal} estimates the children (green) as it should, whereas \textit{Determinstic} does not. \textit{Causal} also maintains the parent (red) as it should whereas \textit{Undirected} changes it. Overall, this demonstrates that \textit{Causal} behaves well in all settings while the other approaches perform only in specific settings, namely when intervening on children for \textit{Deterministic} and on parents for \textit{Undirected}. For our experiments on medical data, we thus seek graphs taking these forms to investigate this effect further.

\subsection{Graph Misspecification}

\begin{figure}[h]
  \begin{minipage}[b]{.25\linewidth}
    \centering
    \includegraphics[width=\linewidth]{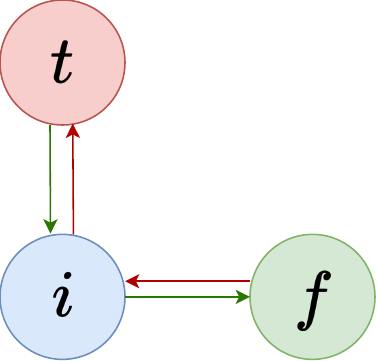}
    \captionof{figure}{}
    \label{fig:misspecified}
  \end{minipage}\hfill
  \begin{minipage}[b]{.7\linewidth}
    \centering
    \scalebox{0.8}{
    \begin{tabular}{c|c|c|c|c}
        Int. & Effect & Det. & Undir. & Causal \\
        \hline
        \multirow{2}{*}{\textcolor{WildStrawberry}{$t$}} & \textcolor{RoyalBlue}{$i$} (\(\uparrow\)) & -6.94 & \textbf{-4.21} & -6.93 \\
        & \textcolor{OliveGreen}{$f$} (\(\uparrow\)) & -4.96 & \textbf{-3.81} & -4.89\\
        \hline
        \multirow{2}{*}{\textcolor{RoyalBlue}{$i$}} & \textcolor{WildStrawberry}{$t$} (\(\uparrow\)) & \textbf{-2.17} & -5.25 & -5.16 \\
        & \textcolor{OliveGreen}{$f$} (\(\uparrow\)) & -5.68 & \textbf{-4.03} & -5.65 \\
        \hline
        \multirow{2}{*}{\textcolor{OliveGreen}{$f$}} & \textcolor{WildStrawberry}{$t$} (\(\uparrow\)) & \textbf{-2.44} & -5.52 & -5.30 \\
        & \textcolor{RoyalBlue}{$i$} (\(\uparrow\)) & \textbf{-3.85} & -6.74 & -6.99 \\
        \hline
    \end{tabular}
    }
    \captionof{table}{}
    \label{tab:misspecified}
  \end{minipage}
  \caption*{Fig 5: True relations (green) are reversed in the misspecified model (red). TABLE V: 
  Similar to Table \ref{tab:ionfg} but for misspecified model (Fig 5). ``Wrong assumptions" affect \textit{Causal}.
  }
\end{figure}

To be beneficial, as in the above analysis, causal inference relies on the assumption that the DAG is correctly specified, capturing the true relationships between the variables. However, in real-world problems, we may not know the true underlying causal relationships. How much can wrong causal assumptions hinder generative models? We explore the ramifications of using an incorrectly specified graph for DAG (b) in Fig.~\ref{fig:mnistdags}. To achieve this, we keep the same data but reverse the assumed direction of the causal relationships in the model (Fig.~\ref{fig:misspecified}). Clearly, this affects only the \textit{Causal} model since the others do not consider the directions of relationships.

As seen in Table \ref{tab:misspecified}, this leads to \textit{Causal} matching the worse of the other two methods in each case. Although this is an extreme case where the arrows are deliberately reversed, it serves as a warning of the importance of using the correct graph if causal models are employed. In real world settings, practitioners may use domain knowledge or statistics from the data to narrow down the space of plausible graphs. Advanced techniques from causal discovery \cite{infdiscovery, discrev} could also be used, though their application to visual data has been limited to date.

\section{Experiments with Chest X-Rays}
\subsection{Setup}
\label{sec:cxr1_setup}

\begin{figure}[h]
  \begin{minipage}[b]{.35\linewidth}
    \centering
    \includegraphics[width=\linewidth]{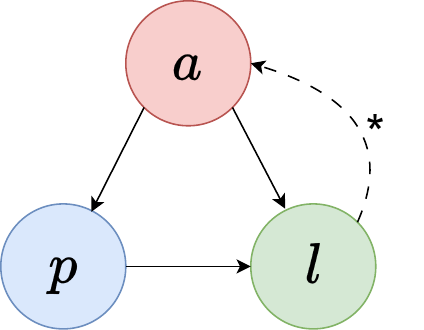}
    \captionof{figure}{}
    \label{fig:chest_dag}
  \end{minipage}\hfill
  \begin{minipage}[b]{.65\linewidth}
    \centering
    \scalebox{0.65}{
    \begin{tabular}{c|c|c|c|c}
        Intervention & Effect (\(\downarrow\)) & Det. & Undir. & Causal \\
        \hline
        \multirow{3}{*}{Age} & Age & 2.983 & 3.176 & \textbf{2.808} \\
        & PE*  & 23.721 & 8.346 &  \textbf{8.335}\\
        & Lungs* & 23.820 & 11.901 &  \textbf{11.713}\\
        \hline
        \multirow{3}{*}{PE} & Age & \textbf{3.455} & 9.711 & 3.788\\
        & PE & 0.066 & 0.069 & \textbf{0.064}\\
        & Lungs*  & 0.476 & \textbf{0.354} & 0.371\\
        \hline
        \multirow{3}{*}{Lung Size} & Age & \textbf{3.700} & 10.640 & 3.825\\
        & PE & 0.318 & 0.564 & \textbf{0.307}\\
        & Lungs & \textbf{0.018} & 0.020 & 0.019 \\
        \hline
    \end{tabular}}
    \captionof{table}{}
    \label{tab:chest}
  \end{minipage}
  \caption*{Fig 6: DAG used for the chest x-ray experiments. Age \((a)\) causes pleural effusion \((p)\), and both cause lung volume \((l)\). TABLE VI: Effect on parent of changing child variable for each of the approaches. Age is measured in MAE, Pleural Effusion (PE) is measured by percentage error. Effects followed by * are evaluated via the proxy approach described above and thus, measured using the intervention.}
\end{figure}

We continue by investigating the generative models on clinical data, using the Chexpert dataset \cite{chexpert} of chest x-rays. We partition the 187k samples at the patient level into a 70/20/10\% train/validation/test split. 
To study causal generative modelling with this data, we identify an appropriate set of variables for which there is clinical evidence of relationships; we posit that \textit{age} causes \textit{pleural effusion} (PE). This association is supported by the data as the mean age of patients with PE is 63.1 while the mean of those without is only 58.7, and a t-test shows that this difference is significant at the 5\% level. Although this does not prove that age causes PE, there is clinical evidence to suggest that this relationship is true \cite{pleuralage}. Regardless, for the purpose of experimenting with causal modelling, we here assume this correlation is causal. It is also generally accepted that PE causes a decrease in \textit{lung volume} as the excess fluid in the pleural space compresses lung tissue \cite{pleuraleffusion}. Finally, there is evidence that lung volume decreases with age \cite{lungage}. These relationships form the DAG in Fig.~\ref{fig:chest_dag}, which we use in our analysis. Finally, we also add \textit{sex} as a conditional variable of the VAE, so that it can be manipulated, although it is not connected to the other graph variables. For each experiment we report the average over 3 seeds.

\subsection{Counterfactuals with Different Conditioning Methods}
\label{subsec:cfs_chexpert}

Using Chexpert, we train models with each of the conditioning approaches. To evaluate the quality of CFs, we use images of the test split and intervene on (change) each of the variables in turn. However, unlike with Morpho-MNIST, here we do not have the ground truth causal effects and data generating process, so we are unable to precisely evaluate whether variables in our image counterfactuals accurately follow the proposed DAG with metrics such as the log-likelihood. Instead, we develop a proxy metric; we first train a separate predictor of each parent \(p\) from its children \(c\), without using the image as input, and denote each respective predictor \(\phi_{c \rightarrow p}\). For evaluation, we intervene on a parent variable \(p\) via \(\text{do}(p\!=\!\tilde{p})\), to obtain the counterfactual image, \(\tilde{x}_{p=\tilde{p}}\). From this, we extract the estimated counterfactual child value, \(\tilde{c} = c(\tilde{x})\) using a separately trained predictor. We then use the trained predictor \(\phi_{c \rightarrow p}\) to estimate the value of the parent post-intervention,

\[\hat{\tilde{p}} = \phi_{c \rightarrow p}(\tilde{c}) = \phi_{c \rightarrow p}(c(\tilde{x}_{p=\tilde{p}}))\]

This allows us to calculate the error \(||\hat{\tilde{p}} - \tilde{p}||\), through which we estimate the effectiveness of the intervention \cite{axiomatic} \footnote{We experimentally verified this metric by testing it on  Morpho-MNIST and comparing it with the ground truth (not shown due to space limitations).}. To obtain a useful predictor \(\phi_{c \rightarrow p}\) when \(c\) is a discrete variable, we train on soft labels, namely the probability output of a predictor for \(c\) when applied on the training images. To ensure that these probabilities are meaningful, we use the validation set for calibration. For parents of variables that are intervened on, we simply check that they are left unchanged, an extension of the Composition metric proposed by \cite{axiomatic}. 

Results from evaluating counterfactuals on the test-split are shown in Table \ref{tab:chest}. These support the findings from Section \ref{sec:mnist} that \textit{Causal} and \textit{Deterministic} perform better at maintaining the value of parent variables when a child variable is changed. This effect is very clear for the effect on age, as these two approaches change it far less than \textit{Undirected}. However, for the effect on the pleural effusion variable, there appears to be greater entanglement as both of these methods flip this variable incorrectly \(\sim \! 30\%\) of the time when lung size is changed. Moreover, the effect on the variable that is intervened on is similar across the models as this does not rely on any variable relationships and is more a measure of the ability of the generative model to respect desired conditions. Overall, the \textit{Causal} approach gives either the best or close-to-best results in all cases, while the other two fail for different types of interventions.

\subsection{Dataset Augmentation}
\label{subsec:cxr2}

We now explore how generating data using these approaches helps ameliorate unwanted biases in training data. We sample the original data in such a way as to induce a strong, spurious correlation between the PE and sex variables. More precisely, we define a ``bias level'' \(b \in [0,1]\). 
For \(b=b_0\), we sample the data so that 
\(b_0\) proportion of male samples are healthy, \(1 - b_0\) have pleural effusion, and the opposite for females ($b_0$ pleural effusion). We then train a PE classifier using this \textit{biased} data before evaluating the predictor on the \textit{unbiased}, original test-split. Such a PE classifier will have sub-optimal performance on unbiased test data, due to using spuriously correlated features to predict PE, such as using obvious anatomical differences between males and females, artificially induced in the training data. 

To study how well they can mitigate the bias, we then train 3 generative models, one per conditioning approach, with the biased training data.
For each of the 3 models separately, we generate images using 2 approaches: counterfactuals and sampling from the latent space (Sec.~\ref{sec:img_generation}). With each approach, we create one synthetic sample per image in the biased training set, with inverse proportion of healthy/PE per sex (inverse bias level, $1\!-\!b$), to rebalance the training data. Finally, for each method, we train a PE classifier on the corresponding augmented, rebalanced training data (synthetics $+$ original biased). We then evaluate the PE classifier on the unbiased test-split, where better generalisation indicates that the corresponding generative method removed the spurious correlations from the data, by better handling of the underlying variables (sex, PE). Results are shown in Table \ref{tab:aug1}.

\begin{table}[h!]
    \centering
    \scalebox{0.8}{
    \begin{tabular}{c|c||c||c|c|c||c|c|c}
         \multirow{2}{*}{\textbf{Labels}} & \multirow{2}{*}{\textbf{Bias}} & \multirow{2}{*}{\textbf{No Aug.}} & \multicolumn{3}{c||}{\textbf{Sampling}} & \multicolumn{3}{c}{\textbf{Counterfactual}}  \\
         \cline{4-9}
         & & & \textbf{Det.} & \textbf{Undir.} & \textbf{Causal} & \textbf{Det.} & \textbf{Undir.} & \textbf{Causal} \\
        \hline
        \multirow{5}{*}{100\%} & 0.5 & 91.4 & \cellcolor{Green!25}\textbf{92.0} & 91.0 & 91.8 & 91.2 & \textbf{91.9} & 91.6 \\
        & 0.8 & 86.4 & \textbf{87.8} & 87.4 & 87.3 & \cellcolor{Green!25}\textbf{88.0} & 87.2 & 87.6 \\
        & 0.9 & 85.1 & 85.1 & \textbf{85.2} & 83.5 & 84.3 & \cellcolor{Green!25}\textbf{86.7} & 85.0 \\
        & 0.95 & 81.8 & 82.2 & \textbf{83.6} & 83.1 & 82.6 & 84.0 & \cellcolor{Green!25}\textbf{84.3} \\
        & 1 & 50.3 & 51.5 & 51.1 & \textbf{51.8} & \cellcolor{Green!25}\textbf{52.8} & 51.3 & 52.7 \\
        \hline
        \multirow{4}{*}{20\%} & 0.5 & 86.2 & 88.3 & 90.3 & \cellcolor{Green!25}\textbf{90.5} & 87.2 & \textbf{88.9} & 88.5 \\
        & 0.8 & 85.0 & 87.2 & 87.1 & \cellcolor{Green!25}\textbf{89.9} & 86.3 & 87.9 & \textbf{89.3} \\
        & 0.9 & 78.7 & 79.6 & \textbf{82.3} & 82.1 & 80.1 & \cellcolor{Green!25}\textbf{82.5} & 82.4\\
        & 0.95 & 77.0 & 77.4 & \cellcolor{Green!25}\textbf{80.7} & 79.8 & 77.4 & 79.5 & \textbf{79.7}\\
        \hline
        \multirow{4}{*}{10\%} & 0.5 & 85.7 & 86.7 & 88.5 & \textbf{88.6} & 87.1 & \cellcolor{Green!25}\textbf{88.7} & 87.9\\
        & 0.8 & 81.5 & 82.4 & \cellcolor{Green!25}\textbf{86.4} & 85.9 & 81.6 & 85.2 & \textbf{86.1}\\
        & 0.9 & 78.7 & 80.9 & \textbf{82.0} & 81.9 & 80.3 & \cellcolor{Green!25}\textbf{82.2} & 82.1 \\
        & 0.95 & 75.5 & 77.4 & \cellcolor{Green!25}\textbf{80.6} & 79.3 & 75.8 & 79.6 & \textbf{79.7}\\
        \hline
        \multirow{4}{*}{5\%} & 0.5 & 85.1 & 86.5 & 88.0 & \cellcolor{Green!25}\textbf{88.5} & 85.9 & \textbf{87.7} & 86.3\\
        & 0.8 & 76.8 & 77.1 & \cellcolor{Green!25}\textbf{81.5} & 81.4 & 75.9 & \textbf{79.1} & 77.8 \\
        & 0.9 & 75.4 & 79.2 & 79.2 & \cellcolor{Green!25}\textbf{80.5} & 76.9 & 78.4 & \textbf{80.1} \\
        & 0.95 & 74.7 & 76.9 & 78.0 & \cellcolor{Green!25}\textbf{79.0} & 75.7 & 76.3 & \textbf{77.5} \\
        \hline
        \multirow{4}{*}{1\%} & 0.5 & 78.3 & 83.5 & 85.0 & \cellcolor{Green!25}\textbf{87.2} & 81.9 & 83.3 & \textbf{83.7}\\
        & 0.8 & 75.6 & 79.3 & \cellcolor{Green!25}\textbf{81.8} & 81.5 & 76.9 & 78.8 & \textbf{80.4}\\
        & 0.9 & 74.7 & 79.3 & \cellcolor{Green!25}\textbf{79.3} & 79.0 & 74.1 & 77.5 & \textbf{77.7}\\
        & 0.95 & 72.8 & 74.4 & 76.1 & \cellcolor{Green!25}\textbf{76.2} & 73.8 & \textbf{75.6} & 74.7\\
        \hline
        \multicolumn{2}{c||}{Average} & 78.90 & 80.65 & 82.15 & \cellcolor{Green!25}\textbf{82.33} & 79.79 & 81.54 & \textbf{81.67} \\
    \end{tabular}}
    \caption{AUC of pleural effusion predictor for raw CheXpert data (no augmentation) and augmented with synthetic data from each of the three conditioning approaches and each of the two generation methods. Green indicates the best AUC in that row. Bold indicates the best performance within the specific generation approach (sampling/counterfactual).}
    \label{tab:aug1}
\end{table}

The results show that augmenting the data with synthetic images improves classifier performance, as shown in the literature when training data for \textit{Deterministic} generative models \cite{googlefair}. Importantly, though, we find this is the case for all labelled percentages and bias levels, suggesting even generative models trained with fewer labels and more biased data can provide benefits. When the dataset is fully labelled, the three conditioning approaches perform similarly. However, as the labelled percentage decreases, we find that \textit{Undirected} and \textit{Causal} tend to outperform \textit{Deterministic}. This suggests that the stronger inductive biases provided by the graph in these models can ameliorate the lack of labels. Since pleural effusion is correlated with a variable (sex) outside of the modelled causal graph (Fig.~\ref{fig:chest_dag}), we hypothesise the additional information provided by the age and lung size variables are key in providing counterracting this bias.

\begin{table}[h]
    \centering
    \begin{tabular}{c|c|c|c|c|c|c}
         \multirow{2}{*}{Metric} & \multicolumn{3}{c|}{Sampling} & \multicolumn{3}{c}{Counterfactual}  \\
         \cline{2-7}
         & Det. & Undir. & Causal & Det. & Undir. & Causal \\
        \hline
        FID & 2.280 & 2.292 & 3.307 & 1.815 & 2.221 & \textbf{1.730} \\
        MMD & 0.054 & 0.049 & 0.055 & 0.051 & 0.048 & \textbf{0.038} \\
        SSIM & 4.37\% & 5.36\% & 4.79\% & 4.01\% & \textbf{3.85\%} & 4.07\%\\
        LPIPS & 0.198 & 0.157 & 0.158 & 0.147 & \textbf{0.135} & 0.150 \\
    \end{tabular}
    \caption{Quality of the data generated by each method.}
    \label{tab:quality}
\end{table}

\begin{figure}[h] 
    \begin{subfigure}[c]{0.5\linewidth}
    \centering
    \includegraphics[width=\textwidth]{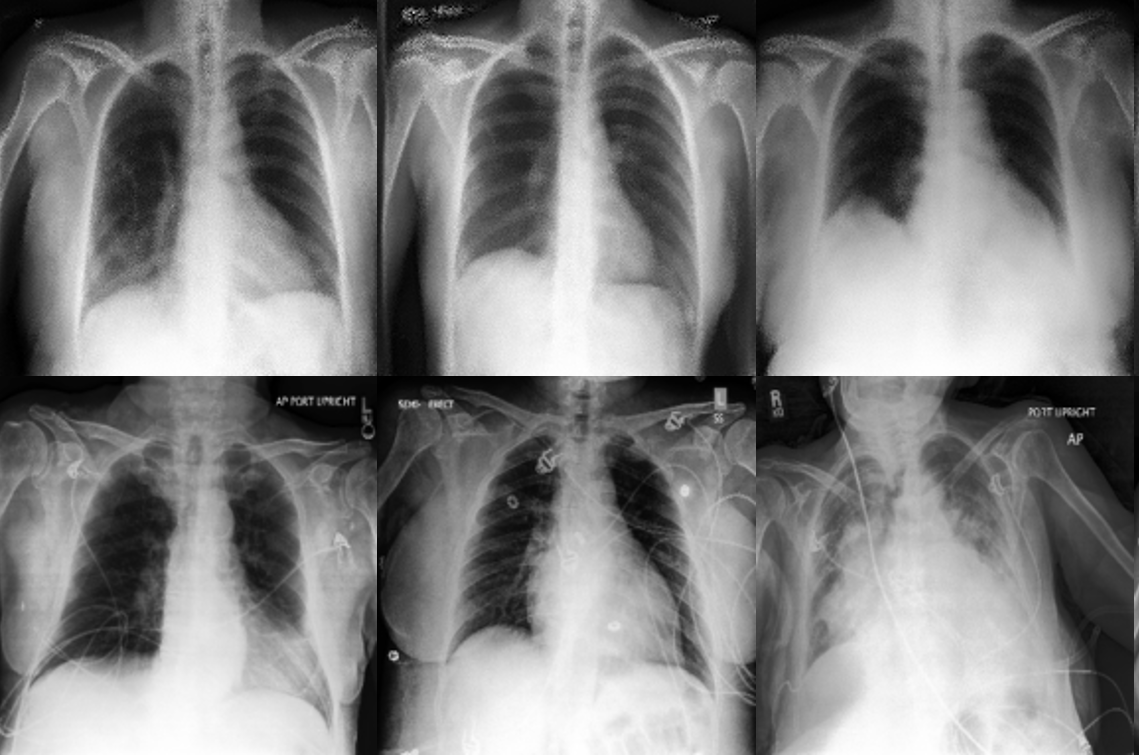}
    \end{subfigure}\hfill
    \begin{subfigure}[c]{0.45\linewidth}
    \centering
    \includegraphics[width=\textwidth]{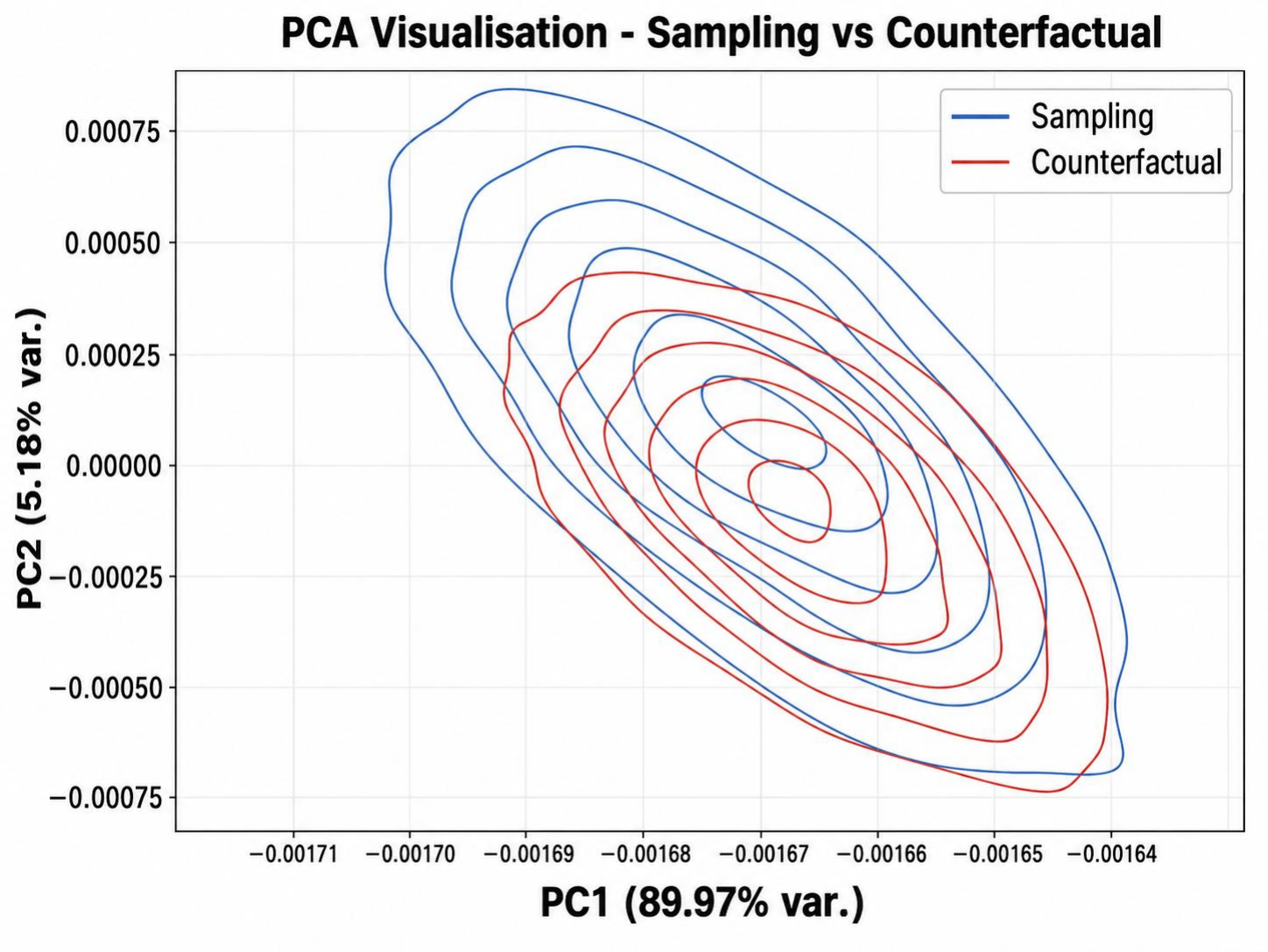}
    \end{subfigure}\hfill
    \caption{Left: Examples of sampled (top) and counterfactual (bottom) images. Right: PCA of embeddings of images generated via Sampling vs Counterfactual when 1\% data is labelled.}
    \label{fig:egspca}
\end{figure}

Another important observation is that for higher labelled percentages, the performance using synthetic data generated via counterfactuals tends to be better than when using sampled data. We believe this is due to the counterfactuals being of slightly better image quality due to the image prior that is provided. To investigate this, we measure the quality of images generated by each method (Table \ref{tab:quality}). For all metrics, we first embed the images to a shared latent space using a DenseNet121 model pre-trained on chest x-ray data \cite{pretrained}. To evaluate image quality, we consider the case with 1\% of the data labelled to show that even in this case, CFs produce better quality, although we find similar results for all label levels. For metrics, we use FID\cite{fid}, maximum mean discrepancy, structural similarity index metric \cite{ssim} and the LPIPS metric \cite{lpips}. This allows us to build a comprehensive picture of how the distribution of generated images compares to our training samples. As shown in Table \ref{tab:quality}, the images produced by the counterfactuals tend to be of higher quality than those produced by sampling. 
Example outputs are shown in Fig.~\ref{fig:egspca} (left), where  the counterfactuals (bottom) appear to be of higher perceptual quality than the sampled images (top).

 \begin{figure*}[h!] 
    \centering
    \includegraphics[width=\textwidth]{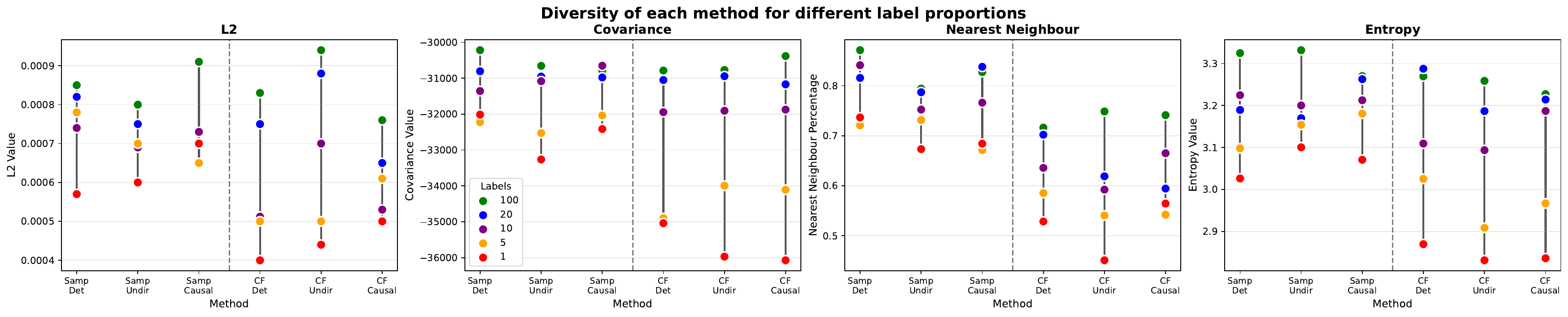}
    \caption{Diversity of the data generated by each of the methods. Higher indicates greater diversity for each metric.}
    \label{fig:diversity}
\end{figure*}

However, in lower label regimes, the sampling approach begins to perform better. We suspect that this is because with a counterfactual we only create a modification of an existing labelled sample. Thus when labelled samples are few, the diversity of counterfactuals is also low. To measure diversity we encode the images into the same latent embedding space and then use a range of metrics such as the average L2 distance between the points in the latent space and the covariance of these points, with larger values for each indicating greater spread and thus, diversity. We also measure the proportion of points with nearest neighbour further than a certain threshold, \(\epsilon = 0.0005\). Finally, we conduct k-means clustering and use these clusters to calculate the entropy of the embedded data. In Fig.~\ref{fig:diversity}, we see that, especially for lower label percentages, sampling results in more diverse images than counterfactuals. This is further portrayed in Fig.~\ref{fig:egspca} (right), which clearly depicts the greater spread of the first two principal components of the embeddings of the images from sampling compared to those from counterfactuals.

Overall, these results suggest that both the \textit{Causal} and \textit{Undirected} approaches provide improved performance over simpler conditioning when augmenting a training dataset. Moreover, when a large proportion of labels are available, the greater realism of counterfactuals provides a benefit over sampling new images. On the other hand, these counterfactuals suffer from a lack of diversity when in low label regimes, when the diversity of the sampled images leads to better downstream results.\\

\subsection{Fairness}
\begin{figure}[h!] 
    \centering
    \includegraphics[width=0.49\textwidth]{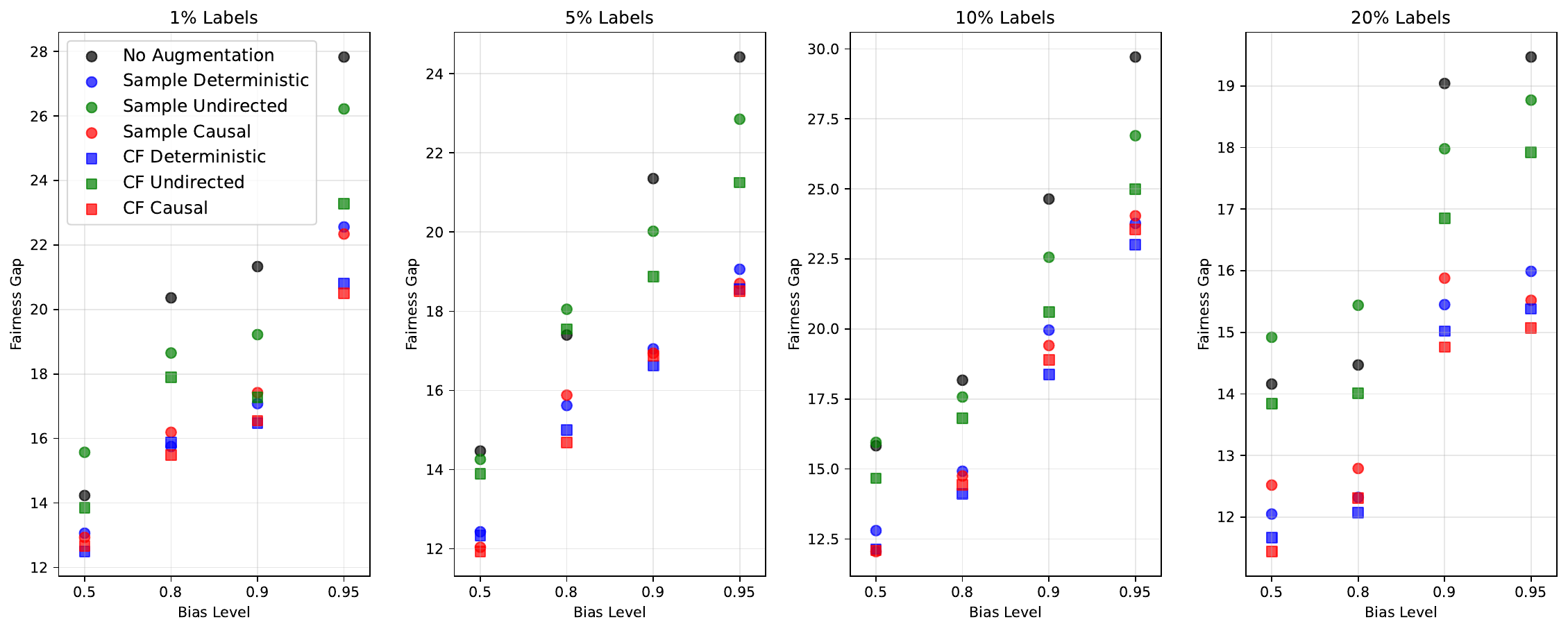}
    \caption{AUC of each method stratified by age group.}
    \label{fig:fairness}
\end{figure}

We extend the previous analysis to explore the effect of the different conditioning approaches on the fairness of downstream predictors. We consider the performance of the PE classifier on five age categories (-20, 20-40, 40-60, 60-80, 80-). We report the difference in AUC between the best and worst performing groups (\textit{fairness gap}) in the same label and bias settings as in Sec.~\ref{subsec:cxr2}, when generating synthetic data such that it counteracts the correlation between PE and sex. Note that even though we consider the fairness gap between age categories, we do not explicitly control for bias in age when augmenting the training data. Results are shown in Fig.~\ref{fig:fairness}. 
The non-augmented data often results in the least fair model, due to the inherent imbalanced prevalence of PE in older groups (Sec.~\ref{sec:cxr1_setup}). Augmentation via the \textit{Undirected} method is the worst at improving fairness with respect to age, which we analyse further below.

We further study fairness under distribution shift. We conduct similar experiments as before, by training generative models on artificially biased CheXpert (10\% labels) and augmenting it with synthetic images, so that the prevalence of PE is balanced between sexes. We then train one PE classifier for each version of the balanced augmented data. Differently than before, however, we now test performance on MIMIC \cite{mimic}, while measuring performance gap with respect to sex, as well as age as before. Table \ref{tab:augmimic} shows the results. We again see that augmentation with synthetic data improves the performance of the PE classifier, in particular the \textit{Undirected} and \textit{Causal} approaches. Moreover, we find that the CFs enhance performance slightly more than unconditional sampling, supporting the previous conclusion from Table \ref{tab:quality}. As we are controlling for sex when synthesising data, we also see that there is a slight improvement (decrease) in the fairness gap for sex with all methods.

 \begin{table}[h]
 \centering
 \scalebox{0.75}{
 \begin{tabular}{|c|l|c|c|c|c|c|}
 \hline
 \multicolumn{2}{|c|}{Method} & Acc. (\(\uparrow\)) & AUC (\(\uparrow\)) & F1 (\(\uparrow\)) & Gap - Sex (\(\downarrow\)) & Gap AUC - Age (\(\downarrow\))\\
 \hline
 \multicolumn{2}{|c|}{No Augmentation} & 0.7262 & 0.8729 & 0.5963 & 0.0214 & 0.0867\\
 \hline
 \parbox[t]{2mm}{\multirow{3}{*}{\rotatebox[origin=c]{90}{Sample}}} & Deterministic & 0.7490 & 0.8918 & 0.6027 & 0.0192 &0.0552 \\
 & Undirected & 0.7638 & 0.8899 & 0.6129 & \textbf{0.0187} & 0.0697\\
 & Causal & 0.7604 & 0.8922 & 0.6183 & 0.0199 &\textbf{0.0519}\\
\hline
\parbox[t]{2mm}{\multirow{3}{*}{\rotatebox[origin=c]{90}{CF}}} & Deterministic & 0.7518 & 0.8905 & 0.6115 & 0.0196 & 0.0622\\
 & Undirected & \textbf{0.7652} & 0.8942 & 0.6186 & 0.0202 & 0.0774 \\
 & Causal & 0.7641 & \textbf{0.8967} & \textbf{0.6207} & 0.0194 & 0.0667\\
 \hline
 \end{tabular}}
 \caption{
 Performance on MIMIC data of methods trained using non-augmented and augmented Chexpert data.
 }
    \label{tab:augmimic}
 \end{table}

However, for age, a variable we do not control for, we find 
there is a statistically significant difference (ANOVA)
between the resultant fairness gaps by the three conditioning approaches. \textit{Deterministic} and \textit{Causal} reduce this gap for both generation methods, whereas \textit{Undirected} provides less benefit. This may be due to the DAG connection from PE to age, which is only present in the \textit{Undirected} construction (in this direction). After intervening on the PE variable for a particular image, such as to synthetise a healthy CF, we then have to sample age conditional on the PE variable. This results in a bimodal distribution of many healthy younger samples and diseased older samples, undesirably over-amplifying the learnt correlation between age and PE in the anti-causal direction.

\begin{figure}[h] 
    \subfloat{
    \includegraphics[width=0.23\textwidth]{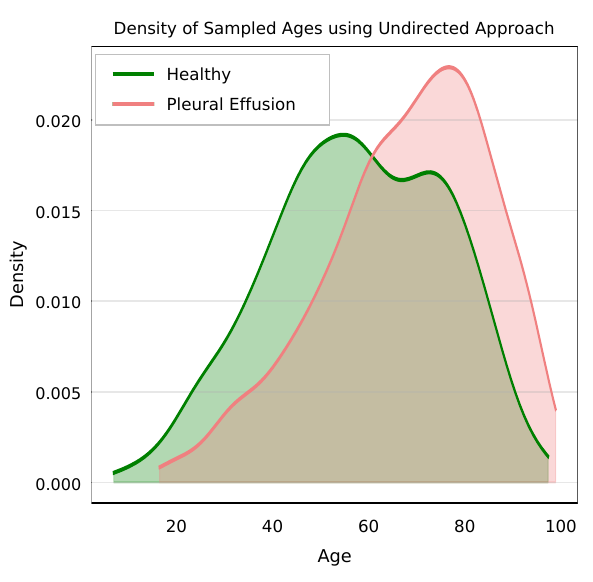}
    }
    \subfloat{
    \includegraphics[width=0.23\textwidth]{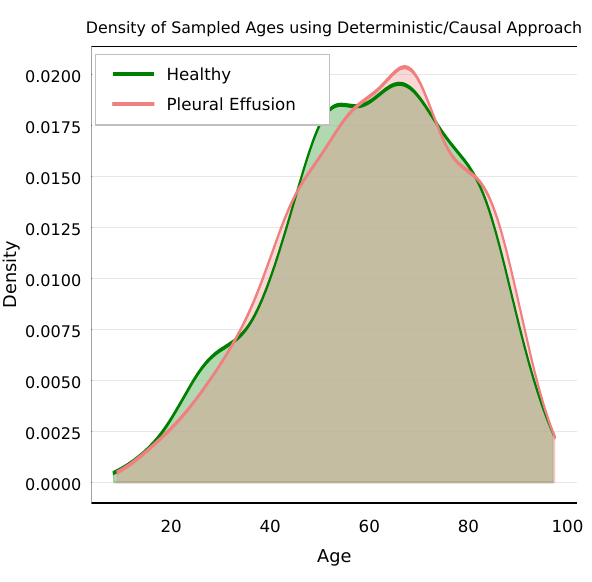}
    }
        \caption{Density estimate of ages from samples produced by each method. \textit{Undirected} samples age depends on disease status, resulting in a difference between ages of samples with pleural effusion (red) and those without (green). \textit{Deterministic} and \textit{Causal} sample age unconditionally, resulting in greater overlap between samples with and without pleural effusion.}
    \label{fig:agehist}
\end{figure}

\begin{figure}[h!] 
    \centering
    \includegraphics[width=0.49\textwidth]{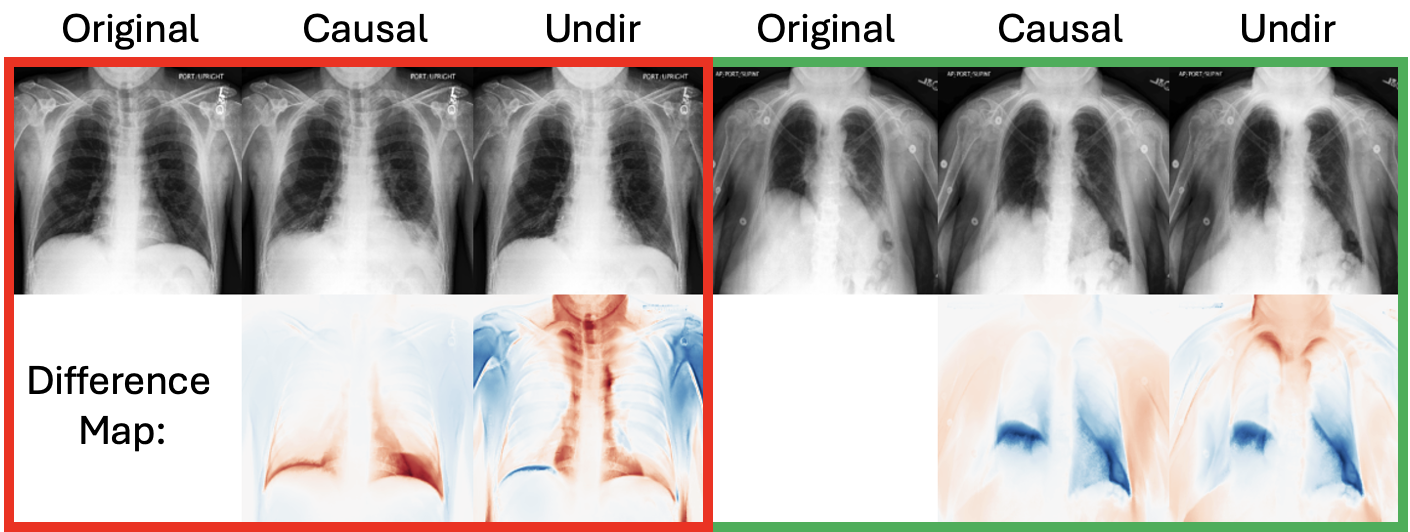}
    \caption{Example interventions using \textit{Causal} vs \textit{Undirected}. On the left (red), we intervene with do(disease) and on the right (green) with do(healthy). For each of these we show the original image, interventions using \textit{Causal} and \textit{Undirected}, and difference maps between original and counterfactuals.}
    \label{fig:egcondvscaus}
\end{figure}

On the other hand, as \textit{Deterministic} and \textit{Causal} do not have this connection (in this direction), age is sampled unconditionally, enabling augmentations that rebalance the existing correlations. These differences are presented visually in the kernel density estimate plots in Fig.~\ref{fig:agehist}. 
Finally, we present two examples of interventions in Fig.~\ref{fig:egcondvscaus} to observe how these differences manifest visually. The \textit{Causal} approach alters almost exclusively the lungs to add/remove PE. Conversely, while \textit{Undirected} changes the lungs, it also alters other parts of the image, potentially because they are affected by variables that are mistakenly changed (e.g. age). In conclusion, these results show that in the setting where we try to mitigate a spurious correlation of 2 variables, here sex and PE, the \textit{Undirected} approach mishandles the parent (age) of the intervened variable (PE). In this setting, \textit{Causal} or \textit{Deterministic} seem preferable.

\section{Conclusion}
In this work, we present and investigated six approaches to generative modelling for synthesising medical images. We delineate three approaches for conditioning a generative model, of increasingly strong assumptions about the causal relationships between variables. 
Moreover, we consider generating images both by synthesising counterfactuals of existing images and synthesising ``de novo" images by sampling from the latent space of the generative model. The study explores which approach is appropriate under different conditions.

Analysis on Morpho-MNIST, where we control the data generating process, shows a clear distinction between the three conditioning approaches for generating counterfactuals. If the correct causal graph has been identified, \textit{Causal} modelling offers consistently high performance on all explored types of DAGs. The \textit{Deterministic} approach is found too rigid and, by design, will only change the variable that we intervene on. Conversely, the \textit{Undirected} approach is too flexible and changes variables more than it should. However, when the graph is misspecified, the advantage of \textit{Causal} modelling disappears. The results highlight that when an ML practitioner is uncertain about the DAG underlying the data generating process, \textit{Deterministic} or \textit{Undirected} conditioning may be preferable over \textit{Causal} modelling.

By experimenting on clinical chest x-rays, although we could not control the data generating process, we obtain key insights about under what conditions causal modelling and counterfactual generation are beneficial. Firstly, the stronger inductive biases in \textit{Causal} and \textit{Undirected} allow these models to better overcome biases and a lack of labels in the data; training on datasets augmented with data generated using these approaches provides improved disease prediction performance over no augmentation and the common \textit{Deterministic} approach. Furthermore, with regards to the generation method, we uncover a trade-off between the quality and diversity of the synthetic data. While sampling new images allows us to generate essentially limitless quantities of data, the counterfactuals are typically more perceptually accurate by being tied to a real image. As such, sampling tends to perform better in low-label settings where the number of possible counterfactuals is limited. Conversely, the latter is favourable in high-label settings where there are more possible counterfactuals and hence, a smaller gap in diversity between approaches. 
Finally, we explore implications of these approaches on fairness. Through experiments on both CheXpert and MIMIC, we discover a latent downside of \textit{Undirected}, namely that the bidirectional relationships may lead it to exacerbate existing biases unless carefully controlled for. This is in contrast to \textit{Deterministic} and \textit{Causal}, provided the proposed DAG is correct. 
Overall, 
we have demonstrated the utility of causal techniques for synthetic data augmentation, under which conditioned each is advantageous, and hope that future research will build on this to design ever more potent models.

\section{Future Work}

This investigation covered three conditioning approaches, using a variant \cite{ibrahim2024semi} of a causal hierarchical VAE \cite{hvae} for its proven capabilities in counterfactual generation. While no assumptions were made about the underlying model, validating our findings using different causal models (e.g. diffusion) would be desirable.
Moreover, we experimented on 2D images, because sufficiently capable 3D causal generative models have not yet been developed, but should they become available, it would be interesting to further validate these insights on volumetric data. This would also allow for richer analysis with data such as brain MRI where there are a greater number of meta-variables and structures within the volume to control. This would lend itself to another natural extension of using a more complex causal graph. However, this also comes with the pitfall that these more subtle changes may not produce visible changes. Furthermore, in this work we have only explored the use of synthetic data to augment training datasets. It would thus be interesting to explore whether the benefits of the causal model in this context carry over to other applications such as stress testing pre-trained models. This is critical for model deployment in real-world clinical settings through both understanding how the model must be improved, and where not possible, how to avoid deploying it in vulnerable settings.

\makeatletter
\renewenvironment{thebibliography}[1]
     {\section*{\refname}%
     \footnotesize
      \list{\@biblabel{\@arabic\c@enumiv}}%
           {\settowidth\labelwidth{\@biblabel{#1}}%
            \leftmargin\labelwidth
            \advance\leftmargin\labelsep
            \usecounter{enumiv}%
            \let\p@enumiv\@empty
            \renewcommand\theenumiv{\@arabic\c@enumiv}}%
      \sloppy
      \clubpenalty4000
      \widowpenalty4000
      \sfcode`\.\@m}
     {\def\@noitemerr
       {\@latex@warning{Empty `thebibliography' environment}}%
      \endlist}
\makeatother

\end{document}